%% file: acl_latex.tex
\documentclass[11pt]{article}

\usepackage[final]{acl}

\usepackage{times}
\usepackage{latexsym}
\usepackage{mdframed}
\usepackage[T1]{fontenc}

\usepackage[utf8]{inputenc}

\usepackage{microtype}

\usepackage{inconsolata}

\usepackage{graphicx}
\usepackage{amsmath}
\usepackage{amssymb}
\usepackage{amsfonts}
\usepackage{float}
\usepackage{hyperref}
\usepackage{url}
\usepackage{xspace}
\usepackage{booktabs}
\usepackage{multirow}
\usepackage{multicol}
\usepackage[ruled,vlined]{algorithm2e}
\usepackage{adjustbox}
\usepackage[most]{tcolorbox}
\usepackage{tcolorbox}
\usepackage{enumitem}
\usepackage{cuted}
\usepackage{mdframed}
\usepackage{array}

\usepackage{tabularx}

\definecolor{CaseTitle}{HTML}{222222}
\definecolor{CaseAccent}{HTML}{1F77B4}    
\definecolor{CaseGood}{HTML}{1A7F37}     
\definecolor{CaseGray}{HTML}{F5F5F5}

\usepackage{listings}
\usepackage{xcolor}

\newcommand{\ourmodel}{SceneAlign\xspace}

\title{SceneAlign: Aligning Multimodal Reasoning to Scene Graphs in Complex Visual Scenes}

\author{
  Chuhan Wang$^{1}\thanks{These authors contributed equally to this work.}$ ,
  Xintong Li$^{1*}$,
  Jennifer Yuntong Zhang$^{2}$,
  Junda Wu$^{1}$,
\\
  \textbf{Chengkai Huang}$^{3}$,
  \textbf{Lina Yao}$^{3}$,
  \textbf{Julian McAuley}$^{1}$,
  \textbf{Jingbo Shang}$^{1}$
\\
  \textsuperscript{1}University of California, San Diego 
  \textsuperscript{2}University of Toronto\\
  \textsuperscript{3}University of New South Wales
\\
  \texttt{\{chw136,xil240,juw069,jmcauley,jshang\}@ucsd.edu} \\
\texttt{jenniferyt.zhang@mail.utoronto.ca}\\
\texttt{\{chengkai.huang1, lina.yao\}@unsw.edu.au} \\
}

\begin{document}
\maketitle

\input{sections/0_abstract}
\input{sections/1_introduction}

\input{sections/2_related}

\input{sections/3_method}

\input{sections/4_exps}
\input{sections/5_case_study}

\section{Conclusion}
We presented \ourmodel, a scene-graph–guided preference alignment framework that improves grounding-faithful reasoning in multimodal large language models. By generating semantically coherent yet structurally inconsistent rationales through controlled scene-graph perturbations, \ourmodel addresses reasoning-grounding inconsistency and explicitly aligns chain-of-thought reasoning with visual structure. Experiments across seven benchmarks show consistent 3–5\% gains on reasoning-intensive tasks, confirming that structure-aware supervision yields significantly more faithful reasoning than traditional text-based perturbations. 

\section*{Limitations}

Our study focuses on single-image reasoning and does not extend to multi-image or video-based inputs, where temporal and cross-view consistency introduce additional grounding challenges.
Moreover, we rely on GPT-based models for generating chains of thought and for automatic evaluation of reasoning coherence and hallucination rates, which may not fully capture fine-grained human judgments. Future work could explore dynamic or temporal scene graphs and adopt human-centered evaluation protocols to provide a more comprehensive assessment of grounding fidelity in open-ended multimodal reasoning.



\bibliography{custom}

\appendix

\input{sections/appendix}

\end{document}

%% file: sections/0_abstract.tex

\begin{abstract}
Multimodal large language models often struggle with faithful reasoning in complex visual scenes, where intricate entities and relations require precise visual grounding at each step. This reasoning unfaithfulness frequently manifests as hallucinated entities, mis-grounded relations, skipped steps, and over-specified reasoning. Existing preference-based approaches, typically relying on textual perturbations or answer-conditioned rationales, fail to address this challenge as they allow models to exploit language priors to bypass visual grounding.
To address this, we propose \ourmodel, a framework that leverages scene graphs as structured visual information to perform controllable structural interventions. By identifying reasoning-critical nodes and perturbing them through four targeted strategies that mimic typical grounding failures, \ourmodel constructs hard negative rationales that remain linguistically plausible but are grounded in inaccurate visual facts. These contrastive pairs are used in Direct Preference Optimization to steer models toward fine-grained, structure-faithful reasoning. Across seven visual reasoning benchmarks, \ourmodel consistently improves answer accuracy and reasoning faithfulness, highlighting the effectiveness of grounding-aware alignment for multimodal reasoning.
\end{abstract}

%% file: sections/1_introduction.tex
\section{Introduction}

\begin{figure*}[t]
\centering
\includegraphics[width=\textwidth]{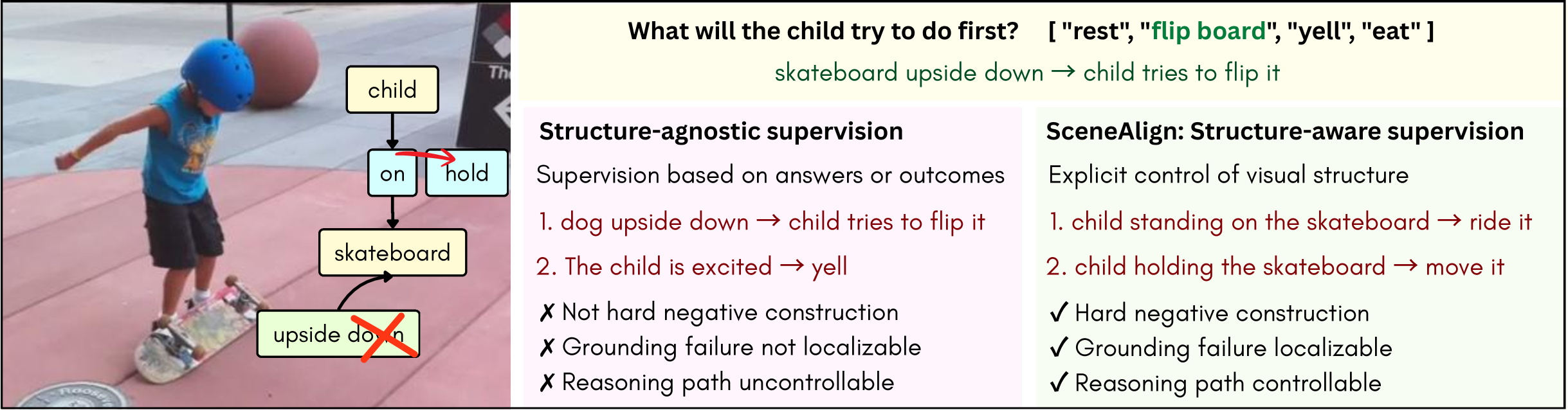}
    \caption{Motivation for structure-aware supervision.
Under structure-agnostic supervision, reasoning errors that ignore, hallucinate, or loosely associate visual structure are indistinguishable at the answer level.
Structure-aware supervision (\ourmodel) perturbs scene-graph elements to generate interpretable and controllable negative examples, making grounding failures observable and localizable.
    }
    \label{fig:motivation-image}
\end{figure*}


Multimodal large language models (MLLMs) have achieved remarkable progress across tasks such as visual question answering, captioning, and instruction following~\cite{yin2024survey,li2024survey}. 
However, they often struggle with visually grounded reasoning, frequently exhibiting reasoning unfaithfulness~\cite{yu2025vfaithlargemultimodalmodels, liu2025thinkingseeingassessingamplified} where the generated rationales are decoupled from the actual visual evidence, even when the final answer appears correct~\cite{wu2025grounded,man2025argus}. This issue becomes particularly severe in complex visual scenes~\cite{akter2024visreas,yang2022modeling,du2023makes,ding2021attention}, where numerous objects and intricate relations require precise, step-wise grounding. 
Typical manifestations of such unfaithfulness include object hallucination, relation mis-grounding (e.g., subject–object swaps), incomplete grounding, and over-specification (e.g., irrelevant or tangential reasoning), all of which disrupt the alignment between textual reasoning and the underlying visual evidence~\cite{chen2024multi}.

Recent studies attempt to mitigate these issues through preference-based alignment of reasoning chains~\cite{ouyang2022training,rafailov2023direct,yu2024rlhf}.
Such methods generate positive and negative rationale pairs by editing tokens or final answers~\cite{zhang2024improve,tan2025answers,zheng2024enhancing,zhou2024can}, encouraging coherent reasoning but leaving visual grounding untouched.
As illustrated in Figure~\ref{fig:motivation-image}, token- or answer-level perturbations (e.g., replacing ``skateboard'' with ``dog'', or generating a negative rationale based on the wrong answer ``yell'') do not create hard, visually grounded negatives, so the model can still rely on language priors without engaging the image. Because these rationales are not tied to specific scene elements, grounding failures cannot be localized or controlled: when the model hallucinates relations, skips steps, or loosely associates objects, we cannot tell which node was misunderstood or where the reasoning chain broke. Thus, realistic grounding errors remain uncaptured.

To address these limitations, we propose \ourmodel, a scene graph–guided preference alignment framework for structure-faithful visual reasoning, where the scene graph represents a set of objects and relations that provide explicit structure to localize reasoning-critical entities.
Given an image and question, \ourmodel first constructs a scene graph that encodes objects, attributes, and relations, and generate a structure-consistent chain of thought (CoT) grounded in this graph.
We then perturb the graph through controlled interventions, such as swapping, removing, or replacing nodes and edges, to simulate realistic grounding failures, and regenerate corresponding CoTs directly from these modified graphs.

For instance, as shown in Figure~\ref{fig:motivation-image}, removing a reasoning-critical node such as ``upside down'' or replacing the relation ``on'' with ``hold'' produces a reasoning chain that remains plausible but omits critical evidence from the scene. 
Guided by the scene graph, such CoTs form high-quality contrastive pairs: visually grounded CoTs as positives, and semantically coherent yet visually inconsistent CoTs as negatives, which expose a diverse set of reasoning failure modes.
Finally, we filter the perturbed samples for plausibility and diversity, and apply Direct Preference Optimization (DPO) over these contrastive pairs to explicitly align the model’s reasoning with the underlying scene structure, enhancing grounding fidelity in complex visual reasoning.

Our contributions are summarized as follows:
\begin{itemize}[leftmargin=*]
    \item We introduce \ourmodel, a scene-graph–guided preference alignment framework for grounding-faithful visual reasoning.
    \item We design controllable perturbation strategies that generate hard and semantically coherent negatives reflecting common grounding failures.
    \item We conduct extensive experiments across multiple visual reasoning benchmarks, showing that \ourmodel consistently improves grounding consistency, reasoning coherence, and answer accuracy over strong baselines.
\end{itemize}

%% file: sections/2_related.tex
\begin{figure*}[htp]
    \centering
    \includegraphics[width=\textwidth]{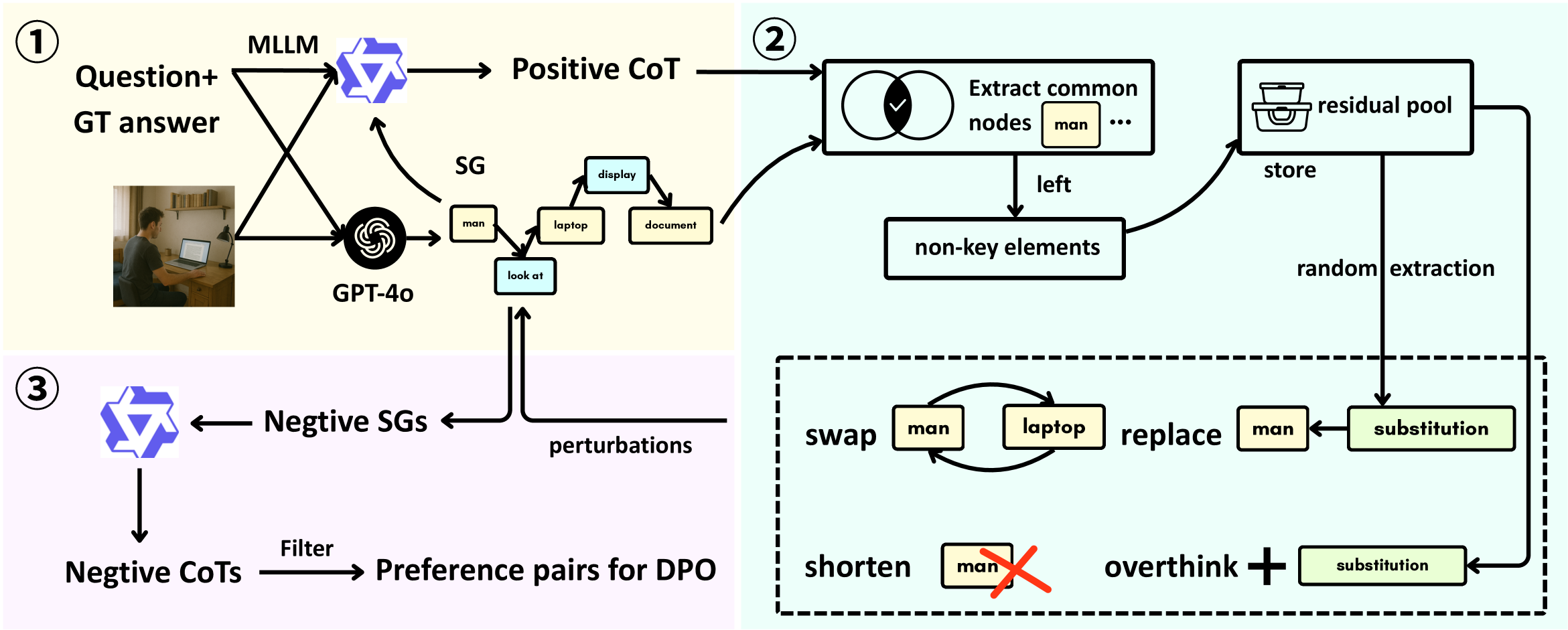}
    \caption{
    \textbf{Framework overview.}
    The \ourmodel pipeline first generates scene-graph–grounded positives (Sec.~\ref{sec:positive}), 
    applies four graph perturbations (\textit{swap, replace, shorten, overthink}) to create negatives (Sec.~\ref{sec:negatives}),
    filters for diversity (Sec.~\ref{sec:select}),
    and finally performs DPO alignment (Sec.~\ref{sec:dpo}).
    }
    \label{fig:image}
\end{figure*}

\section{Related Works}

Recent efforts to improve multimodal reasoning span a broad range of approaches. We specifically build on negative CoT sampling in preference optimization and scene-graph–based multimodal reasoning. \ourmodel perturbs scene graphs to construct structurally inconsistent CoTs and employs preference optimization to align reasoning with the underlying visual structure.

\paragraph{Negative CoT Sampling and Preference Optimization.} 
Modern alignment techniques, such as Reinforcement Learning from Human Feedback (RLHF) \cite{ouyang2022training} and DPO \cite{rafailov2023direct}, have been adapted to multimodal settings to address issues such as hallucination \cite{yu2024rlhf}. A central idea is to generate negative reasoning to provide counterfactual signals. Answer-oriented methods label rationales by final correctness \cite{zhang2024improve, tan2025answers}. However, these approaches often miss structural reasoning errors since flawed reasoning can still produce correct answers. Other approaches perturb CoTs at the token level (e.g., SNSE-CoT \cite{zheng2024enhancing}, QCRD \cite{wang2024qcrd}, and NoRa \cite{zhou2024can}) or treat hallucinated answers as negatives \cite{jiang2024hallucination, sarkar2024mitigating}. 
These approaches demonstrate the value of contrastive supervision, but they rarely capture structural reasoning errors. Recent work instead perturbs relational structures, such as scene graphs for multimodal reasoning \cite{chen2025rrhf}.
We extend prior work through a scene-graph–grounded framework that perturbs entities, attributes, and relations to create counterfactual but semantically plausible rationales, utilizing the structural reasoning that dominates the complex visual scenes.

\paragraph{Scene Graphs for Multimodal Reasoning.} Scene graph integration with large language models has emerged as a promising direction for structured visual understanding. Recent developments in graph-based reasoning reveal the potential for structured representations in complex visual tasks. \cite{mitra2024compositional} utilizes generated scene graphs in zero-shot prompting to extract compositional knowledge without fine-tuning. \cite{lee2024multimodal} extends this approach through multimodal knowledge graphs with relation graph attention networks. Additional methods build on the scene graph CoT \cite{shao2024visual, ji2025enhancing}. Together, they show promise in helping MLLMs to maintain coherence and faithfulness when reasoning about complex scenes by injecting structured representations. Our work extends beyond prior works by integrating scene graphs into the learning process. The MLLM model is trained to prefer rationales that follow the true scene structure over those that deviate from it. Specifically, after generating positive graphs, we perturb them to obtain negative ones, and turn the pairs into reasoning trajectories to train the model for structural faithfulness. This design builds on earlier efforts in scene graph perturbation \cite{singh2023coarse, li2023compositional, huang2024structure}, and leverages these perturbations to align CoT reasoning with preference supervision.

In summary, previous studies have not considered how to align the reasoning process itself with the visual scene structure. They often treat reasoning as a flat textual sequence, neglecting the graph-level dependencies among entities and relations. In contrast, our framework introduces structure-aware perturbations guided by scene graphs, allowing the model to learn from both grounded and structurally inconsistent rationales.


%% file: sections/3_method.tex

\section{Methodology}
To mitigate the inconsistency between answers and their visual grounding, we propose the \ourmodel framework that aligns multimodal reasoning with scene-level structure through preference optimization. 
Figure~\ref{fig:image} provides an overview of our \ourmodel framework. To be specific, given an image-question pair, \ourmodel first constructs scene-graph–grounded positive reasoning, then perturbs scene graphs to produce negative CoTs, filters them for informativeness, and finally applies DPO \cite{rafailov2023direct} to align the model toward structure-consistent reasoning. Algorithm~\ref{alg:scenealign} summarizes its key steps.

\input{images/alg}

\subsection{Positive Reasoning Generation}
\label{sec:positive}
To ensure that the reasoning process is grounded in the visual scene rather than driven by superficial correlations, we first construct \emph{structure-consistent} reasoning traces that explicitly reference scene-graph elements as visual anchors. Formally, a scene graph is defined as,
\begin{equation}
    SG = (E, A, R),
\label{eq:1}
\end{equation}
where $E$ is the set of entities, $A$ the set of attribute pairs, and $R$ the set of relational triples $(e_i, r, e_j)$.

Given a dataset $D$, for each image--question pair $(I,q)\!\in\!D$, we extract a positive scene graph $SG_q^{+}$ that captures all relevant objects, attributes, and relations using GPT\mbox{-}4o~\cite{hurst2024gpt}. A grounded chain-of-thought $\tau_q^{+}\!\in\!\mathcal{T}$ is then generated by a MLLM $\pi_{\theta}$, where $\mathcal{T}$ denotes the space of possible reasoning traces. By anchoring each reasoning step to elements in $SG_q^{+}$, the model is encouraged to reference concrete visual evidence, improving both interpretability and grounding reliability.
This multimodal conditioning ensures that the image remains the primary authoritative signal; any minor noise in the extracted scene graph can be rectified by the MLLM by directly referencing the visual evidence during reasoning.

\subsection{Construction of Negative Scene Graphs}
\label{sec:negatives}

To teach the model what incorrect grounding looks like, we construct structured counterfactuals that reflect common failure patterns, including object hallucination, relation mis-grounding, incomplete grounding, and over-specification, while keeping the overall semantics coherent.
Given a grounded rationale $\tau_q^+$, we first identify its CoT-grounded subgraph
$SG_q^c \subseteq SG_q^+$ that contains only the elements explicitly referenced in
$\tau_q^{+}$. We then construct negative graphs by perturbing elements drawn from
$SG_q^c$ while keeping the remaining part $SG_q^+ \setminus SG_q^c$ unchanged,
ensuring that all modifications remain semantically relevant yet alter the structural
grounding.

Inspired by previous studies on graph-based multimodal alignment \cite{huang2025visual,li2025vision, liu2025aligning}, we introduce four structured transformations: \textbf{swap}, \textbf{replace}, \textbf{shorten}, and \textbf{overthink}, each corresponding to a distinct type of grounding error.
To support replacement-based perturbations, we maintain a \textit{residual pool} $P_{\text{res}} = SG_q^{+} \setminus SG_q^{c}$ consisting of non-key elements from the same scene graph.
Formally, for a relation $(e_i,r,e_j)\in R$ and an element $x\in SG_q^c$,
\begin{itemize}[leftmargin=*]
    \item \textbf{Swap}: exchange the subject and object of a relation,
    \begin{equation}
            {T}_{swap}(e_i,r,e_j) = (e_j,r,e_i).
    \label{eq:2}
    \end{equation}
    This reflects mis-grounding, where subject and object roles are confused, producing fluent rationales that point to wrong bindings.    
    \item \textbf{Replace}: substitute an element using a candidate from the residual pool,
    \begin{equation}
        {T}_{rep}(x) = x', \quad x' \in P_{\text{res}}.
    \label{eq:3}
    \end{equation}
    Such substitutions mimic subtle grounding errors or hallucination-like substitutions, where a non-key scene element replaces a critical one while the rationale remains plausible.
    \item \textbf{Shorten}: remove a key element to simulate omission,
    \begin{equation}
    {T}_{short}(SG_q^c) = SG_q^c \setminus \{x\}, \quad x \in SG_q^c.
    \label{eq:4}
    \end{equation}
    Eliminating key nodes or relations creates traces that skip inference steps, highlighting the need for complete evidence in reasoning chains.
    \item \textbf{Overthink}: augment the graph with redundant or spurious content,
    \begin{equation}
        {T}_{over}(SG_q^{c}) = SG_q^{c} \cup \{x'\}, \quad x' \in P_{\text{res}}.
    \label{eq:5}
    \end{equation}
    Adding redundant grounded details corresponds to over-specification, introducing superfluous steps that distract from the core causal path and reduce reasoning faithfulness.
\end{itemize}
Applying these operators yields a collection of counterfactual scene graphs, 
each constructed by perturbing the CoT-grounded subgraph and reattaching it to the unperturbed remainder,
\begin{equation}
SG^-_{q,i} = \mathcal{O}_i\!\left(SG_q^{c}\right)\oplus\left(SG_q^{+}\setminus SG_q^{c}\right),
\label{eq:6}
\end{equation}
\begin{equation}
\mathcal{O}=\{{T}_{swap}, {T}_{rep}, {T}_{short}, {T}_{over}\}.
\label{eq:7}
\end{equation}
which induces the set,
\begin{equation}
\mathcal{S}_q^- = \{ SG_{q,i}^- \}_{i=1}^k,
\label{eq:8}
\end{equation}
covering the major error types in complex scene reasoning and providing structured negatives for preference optimization.

A \textit{mix} mode randomly samples among perturbations to enhance diversity, while consistency constraints guarantee well-formed outputs. 
This perturbation design allows us to systematically expose recurring error patterns such as hallucination, mis-grounding, skipped inference, and over-specification, so that the model learns to distinguish visual-grounded reasoning from plausible but incorrect alternatives.

\subsection{Selection of Negative Reasoning Chains}\label{sec:select}

The purpose of this step is to construct informative and diverse negative reasoning chains that serve as effective contrastive signals for preference alignment. Without careful selection, negative samples that are either too similar or entirely irrelevant to the positive reasoning chain would provide weak or misleading supervision, undermining the goal of aligning the model’s reasoning with the true scene structure.
Thus, we first perform \textit{scene-graph filtering}. Specifically, 
let
$U(SG)=A\cup\{(e_s,e_o):(e_s,r,e_o)\in R\}$ 
denote the set of attributes and subject–object pairs extracted from a scene graph $SG$, and compute the Jaccard overlap
$J(SG^-_{q,i},SG^+_q)$ between each negative graph and the gold graph,
\begin{equation}
\scalebox{0.98}{$J\!\left(SG^-_{q,i},\,SG^+_q\right)=
\frac{\left|\, U \!\left(SG^-_{q,i}\right)\cap U\!\left(SG^+_q\right)\right|}
{\left|\, U \!\left(SG^-_{q,i}\right)\cup U\!\left(SG^+_q\right)\right|}.$}
\label{eq:9}
\end{equation}
We retain indices
\[
\mathcal{I}_{\mathrm{mid}}
=
\bigl\{
i:
\gamma_\ell
\le
J(SG^-_{q,i},SG^+_q)
\le
\gamma_u
\bigr\},
\]
yielding a filtered set
$\widehat{\mathcal{S}}^-_q=\{SG^-_{q,i}\}_{i\in\mathcal{I}_{\mathrm{mid}}}$.
This filtering yields hard negatives by excluding near-duplicates ($\gamma_u$) and irrelevant outliers ($\gamma_\ell$). Combined with localized perturbations (Sec.~\ref{sec:negatives}), these samples remain semantically consistent with positives yet contain precise grounding errors, forcing DPO to prioritize subtle visual-logical distinctions over global semantic shifts.

Next, we apply \textit{diversity-based sampling}. Let
$z_i=f_\phi(\tau^-_{q,i})\in\mathbb{R}^d$
denote the embedding of the negative CoT generated from $SG^-_{q,i}$.
We then select $m$ diverse negatives via a max–min criterion:
\begin{equation}
\tilde{\mathcal{I}}
=
\arg\max_{\substack{
\mathcal{I}\subseteq\mathcal{I}_{\mathrm{mid}}\\
|\mathcal{I}|=m
}}
\;
\min_{\substack{i\neq j\\ i,j\in\mathcal{I}}}
\|z_i-z_j\|_2.
\label{eq:10}
\end{equation}

This produces the final negative graph set
$\tilde{\mathcal{S}}^-_q=\{SG^-_{q,i}\}_{i\in\tilde{\mathcal{I}}}$
and their associated CoTs
$\tilde{\tau}^-_q=\{\tilde{\tau}^-_{q,i}\}_{i\in\tilde{\mathcal{I}}}$.
By combining filtering with diversity control, we obtain a challenging yet balanced set of counterfactuals that provide rich training signals and prevent overfitting to narrow error modes.

\input{tables/main_result_0}

\subsection{Scene-Graph Preference Optimization}\label{sec:dpo}

To ensure that the model internalizes a preference for scene-faithful reasoning, we optimize it with Direct Preference Optimization (DPO) \cite{rafailov2023direct} using positive and negative chains derived from scene graphs. This encourages the model to assign higher likelihoods to reasoning trajectories that are consistent with the underlying scene structure while penalizing inconsistent ones.

Formally, let $x_q=(I,q,SG^+_q)$ and $\mathcal{P}=\{(x_q,\tau^+_q,\tilde{\tau}^-_{q,i})\}$ be the preference pairs with respect to a reference model $\pi_{\text{ref}}$. DPO optimizes
\begin{equation}
\scalebox{0.7}{$
\begin{aligned}
\mathcal{L}_{\text{DPO}}(\theta)
&= - \frac{1}{|\mathcal{P}|}\sum_{(x,\tau^+,\tau^-)\in\mathcal{P}}
\log \sigma\!\Big(
\beta\big[
\log \pi_\theta(\tau^+\!\mid x)
- \log \pi_\theta(\tau^-\!\mid x) \notag\\
&\quad
- \big(\log \pi_{\text{ref}}(\tau^+\!\mid x)
- \log \pi_{\text{ref}}(\tau^-\!\mid x)\big)
\big]\Big),
\label{eq:11}
\end{aligned}
$}
\end{equation}
where $\sigma$ is the logistic function and $\beta>0$ controls preference sharpness.
The final preference dataset comprises positives $\{\tau^+\}$ and selected negatives $\{\tilde{\tau}^-\}$. DPO trains the MLLM $\pi_\theta$ to prefer reasoning grounded in positive scene graphs, yielding the aligned parameters $\theta^\star$. 
This training objective encourages the model to produce correct answers while maintaining structural faithfulness, thereby improving reasoning across complex scenes.

%% file: images/alg.tex
{\fontfamily{ptm}\selectfont
\begin{algorithm}[t]
\caption{Workflow of \ourmodel}
\label{alg:scenealign}
\fontsize{8.7pt}{11.2pt}\selectfont

\textbf{Preparation:} Parser $\Phi$ (GPT-4o); perturbation operators $\mathcal{O}$ [Eq.~\eqref{eq:7}]; overlap thresholds $[\gamma_\ell, \gamma_u]$.\\
\textbf{Input:} Dataset $D=\{(I,q,a)\}$; base MLLM $\pi_\theta$.\\
\textbf{Output:} Aligned parameters $\theta^\star$.\\[3pt]

Initialize preference set $\mathcal{P} \leftarrow \emptyset$.\\

\ForEach{$(I,q,a)\in D$}{
    // (1) Positive Reasoning Generation (Sec.~\ref{sec:positive})\\
    Extract $SG_q^{+} \leftarrow \Phi(I,q)$; generate $\tau_q^{+} \leftarrow \pi_\theta(I,q,SG_q^{+})$.\\
    
    // (2) Subgraph Extraction (Sec.~\ref{sec:negatives})\\
    Identify $SG_q^{c} \subseteq SG_q^{+}$ referenced in $\tau_q^{+}$.\\
    
    // (3) Negative \textit{SG} Construction (Eqs.~\eqref{eq:2}–\eqref{eq:6})\\
     $\mathcal{S}_q^{-} = \{ \mathcal{O}_i(SG_q^c) \oplus (SG_q^+ \setminus SG_q^c) \mid \mathcal{O}_i \in \mathcal{O} \}$.\\
    
    // (4) Negative CoT Generation\\
    Generate $\{\tau_{q,i}^{-}\}$ by prompting $\pi_\theta$ with $( q, SG_{q,i}^{-})$.\\
    
    // (5) Scene-Graph Filtering (Eq.~\eqref{eq:9})\\
    Compute Jaccard overlap $J(SG_{q,i}^{-}, SG_q^{+})$; retain indices $\mathcal{I}_{\text{mid}}$ where $J \in [\gamma_\ell, \gamma_u]$.\\
    
    // (6) Diversity-based Sampling (Eq.~\eqref{eq:10})\\
    Select $\tilde{\mathcal{I}} = \arg\max_{\mathcal{I} \subseteq \mathcal{I}_{\text{mid}}} \min_{i\neq j}\|z_i-z_j\|_2$; obtain diverse negatives $\{\tilde{\tau}_{q,i}^{-}\}$.\\
    
    // (7) Preference Pair Construction\\
    Define context $x_q = (I, q, SG_q^{+})$; add $\{(x_q, \tau_q^{+}, \tilde{\tau}_{q,i}^{-})\}_{i \in \tilde{\mathcal{I}}}$ to $\mathcal{P}$.\\
}
\vspace{2pt}
// (8) Scene-Graph Preference Optimization (Eq.~\ref{eq:11})\\
$\theta^\star \leftarrow \textsc{DPO-Train}(\pi_\theta, \pi_{\text{ref}}, \mathcal{P})$.\\[2pt]
\textbf{return} $\theta^\star$
\end{algorithm}
}

%% file: tables/main_result_0.tex
\begin{table*}[t]
\centering
\resizebox{\textwidth}{!}{
\begin{tabular}{llcccccccccc}
\toprule
\multirow{2}{*}{\textbf{Model}}  & 
\multirow{2}{*}{\textbf{Method}}  &
\textbf{MME-RW} &
\textbf{EMMA} &
\textbf{ScienceQA} &
\textbf{MMMU} &
\multicolumn{3}{c}{\textbf{Hallusion-Bench}} &
\textbf{GQA} &
\multicolumn{2}{c}{\textbf{SeedBench}} \\
\cmidrule(lr){7-9}
\cmidrule(lr){11-12}
& &
Score ($\uparrow$) &
Score ($\uparrow$) &
Acc. ($\uparrow$)&
Acc. ($\uparrow$)&
aAcc ($\uparrow$)&
fAcc ($\uparrow$)&
qAcc ($\uparrow$)&
Exact ($\uparrow$)&
All ($\uparrow$)&
Img ($\uparrow$)\\
\midrule

\multirow{3}{*}{Qwen2.5-VL-3B}
& Base
& 38.32 & 20.00 & 83.19 & 47.11
& 34.28 & 17.05 & 20.66
& 69.40
& 70.94 & 74.88 \\
& SFT
& 39.81 & 22.75 & 83.26 & 48.09 
& 53.28 & 28.15 & 28.36 & \textbf{69.60} 
& 70.30 & 74.83 \\
& \textbf{SceneAlign}
& \textbf{40.42} & \textbf{23.25} & \textbf{85.10} & \textbf{50.70}
& \textbf{58.25} & \textbf{34.59} & \textbf{33.64}
& \textbf{69.60}
& \textbf{72.09} & \textbf{75.60} \\

\midrule

\multirow{3}{*}{Qwen3-VL-4B}
& Base
& 47.99 & 22.75 & 91.49 & 64.26 
& 57.65 & 34.28 & 34.20 & 62.40 
& 74.80 & 76.93 \\
& SFT
& 47.36 & 24.00 & 91.53 & 66.33 
& 62.10 & 39.45 & 39.20 & \textbf{63.00} 
& 74.55 & 77.12 \\
& \textbf{SceneAlign}
& \textbf{48.86} & \textbf{25.25} & \textbf{92.72} & \textbf{66.67} 
& \textbf{65.35} & \textbf{41.05} & \textbf{41.78} & \textbf{63.00} 
& \textbf{77.03} & \textbf{79.89} \\

\midrule

\multirow{3}{*}{Qwen2.5-VL-7B}
& Base
& 43.98 & 17.75 & 88.71 & 51.11
& 57.31 & 33.24 & 32.09
& \textbf{71.80}
& 74.07 & 77.43 \\
& SFT
& 44.45 & 21.00 & 88.80 & 51.53 
& 60.34 & 36.72 & 35.84 
& 71.40 
& 75.17 & 78.46 \\
& \textbf{SceneAlign}
& \textbf{45.02} & \textbf{22.75} & \textbf{89.12} & \textbf{52.83}
& \textbf{62.88} & \textbf{37.57} & \textbf{38.02}
& 71.40
& \textbf{76.54} & \textbf{79.12} \\

\midrule

\multirow{3}{*}{InternVL3-8B}
& Base
& 49.60 & 21.00 & \textbf{90.77} & 60.51
& 49.58 & 26.03 & 24.45 & 60.20 
& 70.72 & 73.67 \\
& SFT
& 50.33 & 22.25 & 90.02 & \textbf{60.60} 
& 57.84 & 32.96 & 31.42 & 60.40 
& 72.10 & 75.05 \\
& \textbf{SceneAlign}
& \textbf{51.94} & \textbf{23.25} & 90.15 & 60.37 
& \textbf{61.72} & \textbf{36.88} & \textbf{35.01} & \textbf{60.80} 
& \textbf{73.45} & \textbf{76.22} \\

\midrule

\multirow{5}{*}{LLaVA-Next-8B}
& Base
& 37.26 & 15.75 & 80.12 & 40.67
& 26.81 & 18.79 & 12.75
& 73.00
& 57.58 & 72.73 \\
& SFT
& 38.95 & 17.00 & 81.99 & 41.20 
& 39.35 & 20.02 & 14.74 
& 73.00 
& 57.71 & 72.80 \\
& LLaVA-Reasoner
& 37.68 & 14.00 & 81.10 & 42.44
& 41.85 & 20.81 & 12.53
& 72.80
& 58.12 & 73.52 \\
& AoT
& 37.68 & 16.75 & 81.25 & 40.89
& 41.64 & 20.23 & 14.95
& 73.00
& 57.58 & 72.74 \\
& \textbf{SceneAlign}
& \textbf{39.42} & \textbf{19.50} & \textbf{83.25} & \textbf{42.85}
& \textbf{46.89} & \textbf{24.57} & \textbf{18.02}
& \textbf{73.40}
& \textbf{59.80} & \textbf{74.60} \\

\bottomrule
\end{tabular}
}
\caption{Performance on \textbf{reasoning-oriented} benchmarks. 
Higher is better ($\uparrow$). Best results are in bold. Comparison of different methods for constructing CoT-based preference pairs. 
All values are in percentage form.}
\label{tab:reasoning_table}
\vspace{-2mm}
\end{table*}

%% file: sections/4_exps.tex
\section{Experiments}

\subsection{Experimental Settings}
\paragraph{Models.}
We apply SceneAlign to five representative MLLMs: Qwen2.5-VL-3B \cite{bai2025qwen2}, Qwen3-VL-4B\cite{bai2025qwen3vltechnicalreport}, Qwen2.5-VL-7B \cite{bai2025qwen2}, InternVL3-8B\cite{zhu2025internvl3exploringadvancedtraining}, and LLaVA-Next-8B \cite{li2024llava}. 
The three Qwen models primarily differ in scale and vision–language coupling. InternVL3-8B integrates a strong vision encoder with tight cross-modal fusion, while LLaVA-Next-8B combines CLIP with an LLaMA3-8B backbone \cite{dubey2024llama} and serves as a standard baseline.
\paragraph{Training Data.}
We construct training data from A-OKVQA \cite{schwenk2022okvqa} by pairing each image–question instance with a scene-graph–consistent positive CoT and multiple perturbed negatives. A-OKVQA is selected for its challenging open-ended questions requiring both visual grounding and external knowledge, making it a widely adopted benchmark for multimodal reasoning. Negatives are generated through four structured perturbations, then filtered and diversified to form DPO preference pairs. In all experiments, we use 3 negatives per instance as the default setting.

\paragraph{Scene Graph Generation.}
To construct scene graphs for each image-question pair, we use GPT-4o to generate structured scene representations conditioned on both the visual input and the corresponding question. The model is prompted to output entities, attribute pairs, and relational triples in a consistent JSON format, ensuring fine-grained grounding of objects and relations relevant to reasoning. The detailed generation prompt and examples are provided in the Appendix~\ref{sec:prompt_design}.

\paragraph{Baselines.}
We compare SceneAlign against the pretrained model and an SFT baseline (fine-tuned on SceneAlign’s positive CoTs) to isolate the gains from preference alignment.
We also compare SceneAlign with two prior methods, \textbf{AoT}~\cite{tan2025answers} and \textbf{LLaVA-Reasoner}~\cite{zhang2024improve}, for constructing CoT-based preference pairs on A-OKVQA, using \textbf{LLaVA-Next-8B} as the backbone since prior work conducted their evaluations on this model.
AoT reformulates each question–answer pair into an answer-conditioned reasoning task to generate contrastive rationales that logically support or contradict the answer. 
LLaVA-Reasoner adopts a two-stage pipeline that first augments short-answer data with GPT-generated reasoning chains and then constructs outcome-based preference pairs optimized via DPO.
We compare SceneAlign against these approaches as representative \textit{negative CoT construction methods}: AoT and LLaVA-Reasoner generate textual negatives through answer or outcome perturbations, while SceneAlign produces \textbf{structurally perturbed} negatives guided by scene graphs, enabling contrastive alignment at the level of visual grounding.

\input{tables/ablation_step2_full}

\paragraph{Evaluation Benchmarks.}
We evaluate \ourmodel\ on a suite of benchmarks that jointly assess its visual grounding and complex scene reasoning capabilities. For visual grounding, MME-RealWorld~\cite{zhang2024mme} tests perception and grounding under real-world scenes, while GQA~\cite{hudson2019gqa} examines compositional question answering requiring precise grounding of objects and relations. For complex scene reasoning, EMMA-mini~\cite{hao2025can}, SEEDBench~\cite{li2024seed}, HallusionBench~\cite{guan2024hallusionbench}, ScienceQA~\cite{saikh2022scienceqa}, and MMMU-Reasoning~\cite{yue2024mmmu} evaluate multi-step reasoning, hallucination robustness, and structure-dependent inference across multimodal inputs. All evaluations are performed using the LMMs-Eval framework~\cite{zhang2025lmms} for consistency and reproducibility.

\paragraph{Implementation Details.}
To ensure reproducibility, we generated all positive scene graphs using the gpt-4o-2024-11-20 snapshot with a temperature of 0 and fixed random seeds. Training was conducted on 2 NVIDIA A100 GPUs. We set the preference optimization coefficient $\beta=0.1$ and learning rate $5\times10^{-6}$, training for 1 epoch. We employed a per-device batch size of 6 with 20 gradient accumulation steps, resulting in an effective batch size of 120. The maximum input length is 4096 tokens, with prompts truncated to 2048 tokens. All hyperparameters were kept consistent across all models to ensure a fair comparison. Further details on training efficiency are provided in Appendix~\ref{sec:training_curve}.

\input{tables/table_other_methods}
\subsection{Main Result}
Table~\ref{tab:reasoning_table} reports results across reasoning-oriented and vision-centric benchmarks for different MLLMs of varying scales and architectures.
Across all settings, \ourmodel consistently outperforms both the pretrained and SFT baselines, showing that scene graph–guided preference alignment provides benefits beyond positive-CoT supervision, where structure-aware negative samples supply a contrastive grounding signal that discourages incorrect or poorly grounded reasoning paths.
Gains are most pronounced on reasoning-intensive benchmarks such as EMMA, HallusionBench, and MMMU-Reasoning, where \ourmodel yields average improvements of 3\%–5\%. 
These tasks require step-wise compositional reasoning and multi-entity grounding, highlighting that our perturbation strategy exposes fine-grained, structure-level grounding errors that standard textual perturbations overlook. 
Notably, even lightweight models such as Qwen2.5-VL-3B and Qwen3-VL-4B achieve clear gains (e.g., +23.97\% and +7.70\% respectively on HallusionBench), demonstrating that structural grounding is effective even without large capacity. 
Larger models (e.g., Qwen2.5-VL-7B, InternVL3-8B) show similar benefits, implying that \ourmodel provides orthogonal improvements to model capacity. 
On vision-centric benchmarks like GQA, SeedBench, and MME-RealWorld, \ourmodel also delivers consistent gains. We further compare \ourmodel with two prior CoT-based preference construction approaches, LLaVA-Reasoner and AoT, using the LLaVA-Next-8B backbone. 
Results illustrate that \ourmodel surpasses both across nearly all metrics, with particularly large margins on HallusionBench and EMMA, confirming that grounding-aware contrastive pairs provide stronger supervision than text-only perturbations.
Overall, these results demonstrate that \ourmodel is a general and scalable framework for improving multimodal reasoning faithfulness through structure-aware preference alignment.

\subsection{Ablation Studies}

\paragraph{Effect of Perturbation Operators.} Table~\ref{tab:ablation_ops} ablates the four perturbation operators: \emph{swap}, \emph{replace}, \emph{shorten}, and \emph{overthink}. Each operator corresponds to a distinct failure mode (role mis-binding, hallucination, skipped inference, and over-specification). Removing any operator reduces performance on the benchmarks most sensitive to that error type. Full SceneAlign, which integrates all four, achieves the greatest and most balanced improvements.

\paragraph{Effect of Scene Graph Elements.}
Table~\ref{tab:ablation_ops} further ablates the contribution of different scene-graph elements by selectively perturbing entities, relations, and attributes. Relation-level grounding brings the largest performance gains, showing that correctly binding subjects and objects is most critical for grounded reasoning. Entity information also plays a major role, confirming the importance of identifying the right participants. Attribute grounding has a smaller but still consistent effect, suggesting that fine-grained properties provide complementary cues beyond structural role assignment.

\paragraph{Effect of Negative Sampling Strategies.} Table~\ref{tab:ablation_sampling} in the appendix compares three sampling variants: random negatives, diversity-only sampling, and our full method (scene-graph filtering followed by diversity sampling). Random negatives yield weak supervision due to trivial or irrelevant samples, while removing graph filtering introduces overly close or overly distant negatives. Our full method consistently achieves the highest performance, highlighting the value of carefully selecting medium-difficulty negatives.

\begin{figure}[t]
\centering
\includegraphics[width=0.48\textwidth]{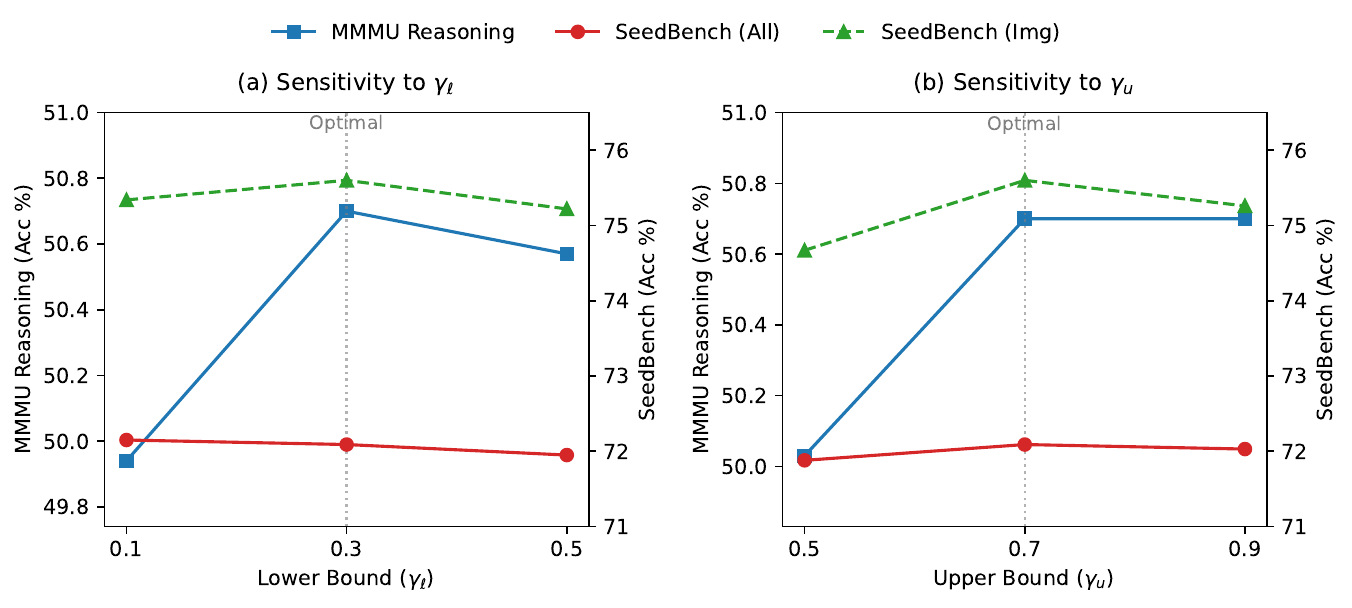}
    \caption{Sensitivity to Jaccard bounds $\gamma_\ell$ and $\gamma_u$, Performance shows an inverted U-shape with optimal performance at $\gamma_\ell = 0.3$ and $\gamma_u = 0.7$.}
    \label{fig:sensitivity_analysis}
\end{figure}

\paragraph{Effect of Overlap Thresholds.}
To study how scene-graph overlap thresholds affect preference construction, we vary the lower bound $\tau_\ell$ and upper bound $\tau_u$. Overly narrow ranges filter out many candidate pairs, reducing data coverage and causing moderate accuracy drops, while overly loose ranges introduce trivial or overly distant negatives that weaken supervision. We find that moderate thresholds strike the best balance; the default SceneAlign setting ($\tau_\ell = 0.3$, $\tau_u = 0.7$) consistently yields strong and well-balanced performance on MMMU-Reasoning and SEED-Bench (Figure \ref{fig:sensitivity_analysis}).

\paragraph{Effect of the Number of Negatives.} Table~\ref{tab:ablation_sampling} studies the impact of using different numbers of negatives per instance. One negative provides limited coverage of reasoning errors, while three negatives significantly improve structural supervision with modest additional cost. Larger numbers offer diminishing returns. We adopt three negatives per instance by default as the best trade-off between efficiency and effectiveness.
\input{tables/ablation_numbers}
\input{tables/ablation_new}

%% file: tables/ablation_step2_full.tex
\begin{table*}[t]
\centering
\resizebox{\textwidth}{!}{
\begin{tabular}{lcccccccccc}
\toprule
\multirow{2}{*}{\textbf{Method}} &
\textbf{MME-RW} &
\textbf{EMMA} &
\textbf{ScienceQA} &
\textbf{MMMU} &
\multicolumn{3}{c}{\textbf{Hallusion-Bench}} &
\textbf{GQA} &
\multicolumn{2}{c}{\textbf{SeedBench}} \\
\cmidrule(lr){6-8}
\cmidrule(lr){10-11}
&
Score ($\uparrow$) &
Score ($\uparrow$) &
Acc. ($\uparrow$)&
Acc. ($\uparrow$)&
aAcc ($\uparrow$)&
fAcc ($\uparrow$)&
qAcc ($\uparrow$)&
Exact ($\uparrow$)&
All ($\uparrow$)&
Img ($\uparrow$)\\
\midrule

Base
& 38.32 & 20.00 & 83.19 & 47.11
& 34.28 & 17.05 & 20.66
& 69.40
& 70.94 & 74.88 \\

\midrule

w/o swap
& 40.13 & 19.25 & 83.55 & 47.60
& 57.52 & 33.82 & 32.09
& 69.40
& 71.30 & 75.35 \\

w/o replace
& 39.60 & 19.00 & 83.70 & 49.65
& 57.20 & 33.24 & 32.53
& 69.60
& 71.65 & 75.50 \\

w/o shorten
& 39.86 & \textbf{23.25} & 84.00 & 48.85
& 58.04 & 34.39 & 33.41
& 69.60
& 71.80 & 75.65 \\

w/o overthink
& 39.55 & 20.75 & 83.45 & 47.90
& 57.75 & 34.02 & 33.23
& 69.20
& 72.05 & \textbf{75.75} \\

\midrule

entity
& 39.76 & 22.75 & 84.31 & 47.57
& 55.84 & 30.90 & 30.13
& 69.40
& 71.48 & 75.42 \\

relations
& 39.76 & 21.25 & 83.19 & 49.00
& 56.62 & 32.35 & 29.91
& \textbf{69.60}
& \textbf{72.16} & \textbf{75.75} \\

attributes
& 39.92 & 19.50 & 83.35 & 46.92
& 55.42 & 32.35 & 29.69
& 69.40
& 71.72 & 75.58 \\

\midrule

\textbf{SceneAlign}
& \textbf{40.42} & \textbf{23.25} & \textbf{85.10} & \textbf{50.70}
& \textbf{58.25} & \textbf{34.59} & \textbf{33.64}
& \textbf{69.60}
& 72.09 & 75.60 \\

\bottomrule
\end{tabular}
}
\caption{Ablation study of structured perturbations on reasoning-oriented benchmarks using \textbf{Qwen2.5-VL-3B}. 
The relative performance order of each variant across benchmarks is preserved, with consistent overall growth. 
All values are in percentage form. Higher is better ($\uparrow$). Best results are in bold.}
\label{tab:ablation_ops}
\vspace{2mm}
\end{table*}

%% file: tables/table_other_methods.tex

%% file: sections/5_case_study.tex
\subsection{Case Study}

To better illustrate how scene-graph grounding improves reasoning faithfulness, 
we present a case study in Figure \ref{fig:case}. The example asks: \textit{“What kind of activity with respect to the motorcycle is the man on the floor most likely engaging in?”} 
The scene contains several interacting entities, including the man, the motorcycle, and a piece of paper, which calls for structured visual reasoning that goes beyond surface-level cues.

\paragraph{Positive CoT.}
Our model correctly grounds each reasoning step in the scene graph: the man is on the ground next to the motorcycle, others are gathered around, and one holds a paper while looking toward the motorcycle, which jointly supports \textit{inspecting or diagnosing} the motorcycle.

\paragraph{Prior Methods.}
Answer-based approaches fix an incorrect final answer (e.g., \textit{riding}) and fabricate reasoning to justify it, leading to fluent but semantically inconsistent CoTs. Token-level perturbations merely alter surface text and often yield logically broken scenes (e.g., ``arguing with the tree''), which is grammatically correct yet structurally invalid.

\paragraph{\ourmodel (Ours).}
In contrast, \ourmodel perturb the \textbf{scene graph structure} through four controlled operations: \textit{swap}, \textit{replace}, \textit{shorten}, and \textit{overthink}. 
Each produces a natural yet structurally inconsistent CoT, revealing distinct failure modes such as 
role misalignment, hallucination, missing evidence, and over-specification. 
These structured negatives expose reasoning drift that purely text-based perturbations fail to capture. Full CoT details are provided in Appendix~\ref{sec:case_study_explanation}.

%% file: sections/appendix.tex
\newpage

\section{Appendix}
\label{sec:appendix}

\subsection{Analysis of Data Reliability}
To validate the quality of our structural interventions and positive rationales, we conducted a human spot check on 150 randomly sampled A-OKVQA instances. Our evaluation confirms that the gpt-4o-generated scene graphs correctly capture all question-relevant entities and relations in all of the cases. 
While occasional redundant nodes (e.g., background objects) appear, they do not degrade the quality of the positive CoTs, 
as the MLLM conditions primarily on the \textit{(image, question)} pair and treats the scene graph as auxiliary guidance. This effectively bypasses extraction noise and underscores \ourmodel’s high-fidelity data synthesis.

\subsection{Prompt Design}
\label{sec:prompt_design}
We use structured prompts (Figure \ref{prompt:p1}, \ref{prompt:p2}, \ref{prompt:p3}) to decouple scene grounding from reasoning. A scene-graph prompt first extracts entities, attributes, and relations in strict JSON format to anchor reasoning to visual structure. We then generate positive CoTs using both the image and graph, and negative CoTs using only the graph, enabling controlled contrast in structure consistency.

\subsection{Case Study Explanation}
\label{sec:case_study_explanation}
We present a case study (Table \ref{tab:case_study_scenealign}) showing that SceneAlign yields structure-consistent positive CoTs, while perturbations expose distinct failure modes.

\subsection{Training Curves}
\label{sec:training_curve}
\begin{figure*}[h]
    \centering
    \includegraphics[width=\textwidth]{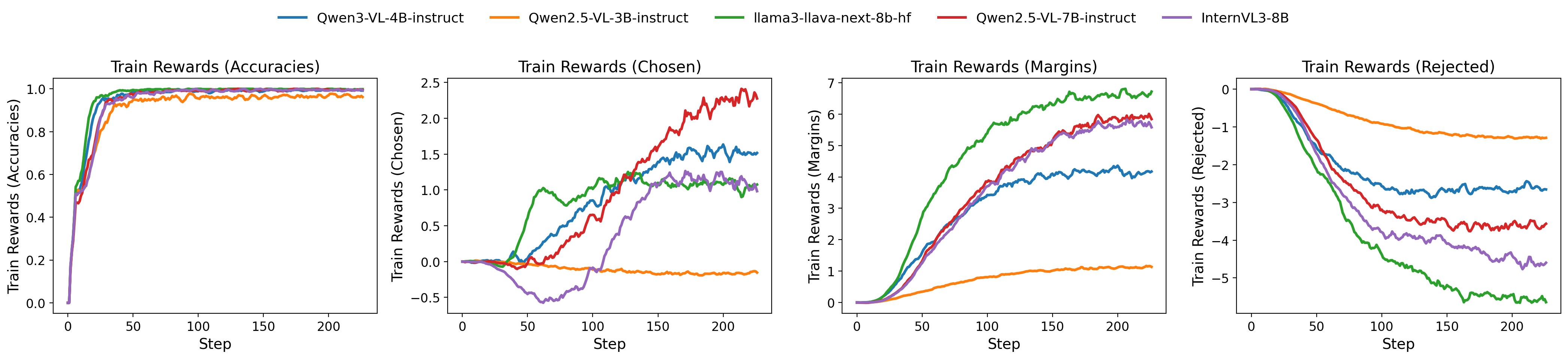}
    \caption{
   Training curves of five multimodal models (Qwen2.5-VL-3B-Instruct, Qwen3-VL-4B-Instruct, Qwen2.5-VL-7B-Instruct, InternVL3-8B, and Llama3-LLaVA-Next-8B-hf) across four reward-related metrics: accuracies, chosen, margins, and rejected, evaluated over 1 epoch.
    }
    \label{fig:plot_11}
\end{figure*}

We visualize four reward-related training curves (Figure \ref{fig:plot_11}) for five MLLMs to validate our method.

\subsection{LLM Usage} Large language models were used exclusively to improve grammar and make minor wording edits in parts of this paper.
\begin{figure}[b]
    \centering
    \includegraphics[width=0.75\linewidth,trim=0 80 680 0,clip]{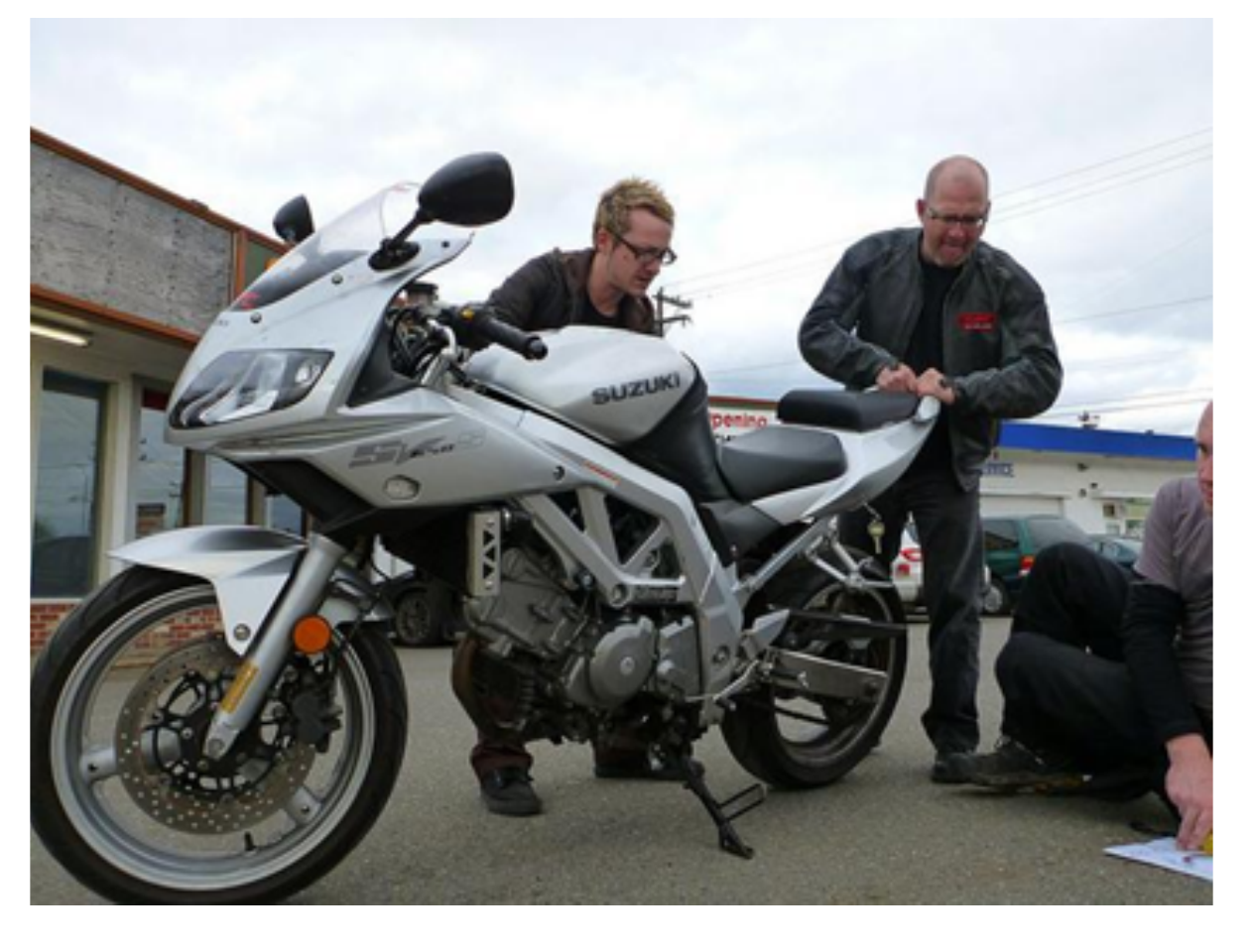}
    \caption{Example Image from Case Study.
    }
    \label{fig:case}
    \vspace{-2pt}
\end{figure}

\begin{figure*}[b]
\centering
\begin{tcolorbox}[
    colback=green!1,         
    colframe=black,  
    arc=6pt,
    boxrule=0.8pt,
    width=\linewidth,
    title=\textbf{Structured Scene Graph Generation Prompt},
    coltitle=black,           
    colbacktitle=green!75!blue!30, 
    fonttitle=\bfseries,
    enhanced
]

\textbf{You are given an image and its associated question.  
Your task is to generate a scene graph in \emph{strict JSON format} that includes the following three fields:}\\
    1. \texttt{"entity"}: a list of all objects and concepts relevant to answering the question.\\
    2. \texttt{"attribute pairs"}: a list of \texttt{[object, attribute]} pairs describing each entity’s key features (e.g., color, size, state).\\
    3. \texttt{"relationships"}: a list of \texttt{[subject, relation, object]} triples describing spatial or semantic relationships.

\vspace{0.5em}
\textbf{Format Example:}

\begin{lstlisting}[
    basicstyle=\ttfamily\small,
    frame=single,
    rulecolor=\color{black!20},
    backgroundcolor=\color{white}
]
{
  "entity": ["man", "motorcycle", "paper", "ground"],
  "attribute pairs": [
    ["motorcycle", "silver"],
    ["paper", "white"],
    ["ground", "paved"]
  ],
  "relationships": [
    ["man", "look at", "motorcycle"],
    ["man", "crouch on", "ground"],
    ["man", "hold", "paper"],
    ["motorcycle", "stand on", "ground"]
  ]
}
\end{lstlisting}

\vspace{0.5em}
\textbf{Attention:}\\
    1. Only return a valid JSON object with the three required fields.\\
    2. Do \textbf{not} include any explanations or natural language text.\\
    3. Ensure the format strictly matches the example above.
\vspace{0.5em}

\textbf{Question:} \texttt{\{question\}}, \texttt{\{ground-truth answer\}}
\vspace{0.5em}
\\
\textbf{Scene Graph:}
\end{tcolorbox}
\caption{Prompt for structured scene graph generation.}
\label{prompt:p1}
\end{figure*}

\begin{figure*}[h]
\centering
\begin{tcolorbox}[
    colback=green!1,         
    colframe=black,  
    arc=6pt,
    boxrule=0.8pt,
    width=\linewidth,
    title=\textbf{Positive CoT Generation Prompt},
    coltitle=black,           
    colbacktitle=green!75!blue!30, 
    fonttitle=\bfseries,
    enhanced
]

\textbf{You are given a scene graph and its associated question and image.  
Your task is to provide step-by-step reasoning to answer the question based on the image and scene graph.
Do not mention the data source.
Treat the scene graph elements as the visual scene itself.}
\vspace{0.5em}

\textbf{Format Example:}

\begin{lstlisting}[
    basicstyle=\ttfamily\small,
    frame=single,
    rulecolor=\color{black!20},
    backgroundcolor=\color{white}
]
1. ...
2. ...
3. ...
4. ...
Conclusion: ...

\end{lstlisting}
\vspace{0.5em}

\textbf{Scene Graph:} \texttt{\{scene\_graph\}}

\vspace{0.5em}
\textbf{Question:} \texttt{\{question\}}, \texttt{\{ground-truth answer\}}

\vspace{0.5em}
\textbf{Step-by-step reasoning:}
\end{tcolorbox}
\caption{Prompt for generating positive chain-of-thought reasoning from image and its corresponding scene graph.}
\label{prompt:p2}
\end{figure*}

\begin{figure*}[h]
\centering
\begin{tcolorbox}[
    colback=green!1,         
    colframe=black,  
    arc=6pt,
    boxrule=0.8pt,
    width=\linewidth,
    title=\textbf{Negative CoT Generation Prompt},
    coltitle=black,           
    colbacktitle=green!75!blue!30, 
    fonttitle=\bfseries,
    enhanced
]

\textbf{You are given a scene graph and its associated question.  
Your task is to provide step-by-step reasoning to answer the question based on the scene graph.
Do not mention the data source.
Treat the scene graph elements as the visual scene itself.}

\vspace{0.5em}

\textbf{Format Example:}

\begin{lstlisting}[
    basicstyle=\ttfamily\small,
    frame=single,
    rulecolor=\color{black!20},
    backgroundcolor=\color{white}
]
1. ...
2. ...
3. ...
4. ...
Conclusion: ...

\end{lstlisting}
\vspace{0.5em}

\textbf{Scene Graph:} \texttt{\{scene\_graph\}}
\vspace{0.5em}

\textbf{Question:} \texttt{\{question\}}

\vspace{0.5em}
\textbf{Step-by-step reasoning:}
\end{tcolorbox}
\caption{Prompt for generating negative chain-of-thought reasoning from a structured scene graph.}
\label{prompt:p3}
\end{figure*}

\input{tables/ablation_step4}

\input{tables/case_study_full}

%% file: tables/ablation_step4.tex
\begin{table*}[h]
\centering
\resizebox{\textwidth}{!}{
\begin{tabular}{lcccccccccc}
\toprule
\multirow{2}{*}{\textbf{Method}}  &
\textbf{MME-RW} &
\textbf{EMMA} &
\textbf{ScienceQA} &
\textbf{MMMU} &
\multicolumn{3}{c}{\textbf{Hallusion-Bench}} &
\textbf{GQA} &
\multicolumn{2}{c}{\textbf{SeedBench}} \\
\cmidrule(lr){6-8}
\cmidrule(lr){10-11}
&
Score ($\uparrow$) &
Score ($\uparrow$) &
Acc. ($\uparrow$)&
Acc. ($\uparrow$)&
aAcc ($\uparrow$)&
fAcc ($\uparrow$)&
qAcc ($\uparrow$)&
Exact ($\uparrow$)&
All ($\uparrow$)&
Img ($\uparrow$)\\
\midrule

Base
& 43.98 & 17.75 & 88.71 & 51.11
& 57.31 & 33.24 & 32.09
& \textbf{71.80}
& 74.07 & 77.43 \\

Random
& 44.22 & 19.25 & 88.79 & 51.48
& 58.15 & 32.08 & 33.19
& 71.40
& 74.25 & 77.60 \\

w/o scene-graph filter
& 44.68 & 22.00 & 88.83 & 51.05
& 62.15 & 36.99 & 38.02
& 71.60
& 75.02 & 78.10 \\

\midrule

num=1
& 44.29 & 20.00 & 88.80 & 51.52
& 58.15 & 32.65 & 32.97
& \textbf{71.80}
& 74.45 & 77.72 \\

num=2
& 44.97 & 21.50 & 88.85 & 51.86
& 61.83 & 36.13 & 37.36
& 71.00
& 75.38 & 78.26 \\

num=4
& 44.89 & 22.25 & 88.91 & 52.24
& \textbf{62.88} & 36.94 & 36.76
& 71.60
& 75.95 & 78.64 \\

\midrule

\textbf{SceneAlign} (num=3)
& \textbf{45.02} & \textbf{22.75} & \textbf{89.12} & \textbf{52.83}
& \textbf{62.88} & \textbf{37.57} & \textbf{38.02}
& 71.40
& \textbf{76.54} & \textbf{79.12} \\

\bottomrule
\end{tabular}
}
\caption{Ablation on negative CoT sampling strategies using \textbf{Qwen2.5-VL-7B}. All values are in percentage form. Higher is better ($\uparrow$). Best results are in bold. }
\label{tab:ablation_sampling}
\end{table*}

%% file: tables/case_study_full.tex
\begin{table*}[h] 
\centering 
\scriptsize
\begin{tabularx}{\linewidth}{@{}p{3.6cm} X@{}} 
\toprule
\textbf{Case Study: Scene-Graph--Grounded Reasoning} 
& \textbf{Question:} What kind of activity with respect to the motorcycle is the man on the floor most likely engaging in? \\[2pt]
\midrule
\textbf{Image} 
& Figure \ref{fig:case}\\
\midrule

\textbf{Ground Truth / MLLM Answer} 
& \textbf{Ground Truth:} Inspecting / Diagnosing \\
& \textbf{MLLM Answer:} Inspecting. \textcolor{green!50!black}{(correct)} \\[4pt]
\midrule
\textbf{Scene Graph} &
\parbox[t]{\linewidth}{%
  \raggedright\ttfamily\footnotesize
  \{ "entity": ["man", "motorcycle", "ground", "paper", "building", "window", "car"],\\
  \quad "attribute pairs": [
    ["motorcycle", "silver"],
    ["motorcycle", "parked"],
    ["ground", "paved"],
    ["paper", "white"],
    ["building", "white"],
    ["window", "glass"],
    ["car", "parked"]
  ],\\
  \quad "relationships": [
    ["man", "look at", "motorcycle"],
    ["man", "crouch on", "ground"],
    ["man", "hold", "paper"],
    ["motorcycle", "stand on", "ground"],
    ["building", "behind", "motorcycle"],
    ["car", "behind", "motorcycle"]
  ]\}\\
}
\\[4pt]

\midrule

\textbf{CoT-grounded Subgraph} &
\parbox[t]{\linewidth}{%
  \raggedright\ttfamily\footnotesize
  \{ "entity": ["man","motorcycle","paper","ground"],\\
  \quad "attribute pairs": [["motorcycle","silver"],["motorcycle", "parked"],["paper","white"],["ground","paved"]],\\
  \quad "triples": [["man","look at","motorcycle"],["man","crouch on","ground"],\\
  \quad\quad ["man","hold","paper"],["motorcycle","stand on","ground"]] \}
}
\\[4pt]

\midrule

\textbf{Positive CoT} 
& 1. The man is on the floor next to the motorcycle, not sitting on it. \\
& 2. Several people are gathered around the motorcycle, suggesting a repair or inspection setting. \\
& 3. One person holds a piece of paper, likely a manual or document. \\
& 4. The man’s position implies he is interacting with the motorcycle. \\
& \textbf{Conclusion:} He is most likely inspecting or performing maintenance on the motorcycle. \\[4pt]
\midrule

\textbf{Prior: Answer-Driven} 
& 1. The question already frames the situation around a motorcycle, which naturally suggests a riding scenario.\\
& 2. In everyday situations, people who interact with motorcycles are most often using them rather than performing technical diagnostics.\\
& 3. Without explicit evidence of a repair task, it is reasonable to treat the activity as related to normal use.\\
& \textbf{Conclusion:} So the man is more likely preparing for or engaging in a riding activity, instead of diagnosing the motorcycle.\\[4pt]

\midrule
\textbf{Prior: Token-level perturb} 
& 1. The man is on the floor next to the banana, not sitting or riding on it.\\
& 2. Several people are gathered around the sky, suggesting a repair or inspection of clouds.\\
& 3. One person holds a dog, likely a manual or document.\\
& 4. The man’s position and the scene imply he is arguing with the tree.\\
& \textbf{Conclusion:} He is most likely cooking dinner, not maintaining the motorcycle.\\[4pt]

\midrule

\textbf{\ourmodel swap} 
& \textit{Change:} ["man","look at","motorcycle"] → ["motorcycle","look at","man"] \\ 
& 1. A motorcycle looking at the man, instead of the man looking at it.\\
& 2. Since the motorcycle is described as the active observer, the man is no longer directly interacting with it.\\
& 3. The man is sitting on the paved ground, holding a piece of white paper.\\
& 4. Without a clear connection between the man and the motorcycle, his activity is interpreted as passive or unrelated to the bike.\\
& \textbf{Conclusion:} The man is most likely just sitting and reading the paper, not doing anything with the motorcycle.\\[4pt]

\midrule

\textbf{\ourmodel replace} 
& \textit{Change:} Replace the entity “paper” with a randomly selected entity from the residual pool (“window”).\\
& 1. The man is crouching on the paved ground near a silver motorcycle.\\
& 2. He is described as holding a white window in his hands.\\
& 3. The man is looking toward the motorcycle while holding the window, but there is no clear indication that he is using the window for repair or maintenance.\\
& 4. Instead, the window appears to be an unrelated object that he is simply handling while being near the motorcycle.\\
& \textbf{Conclusion:} The man is most likely handling or examining the window, rather than engaging in an activity related to the motorcycle.\\[4pt]
\midrule

\textbf{\ourmodel shorten} 
& \textit{Change:} delete a main entity "man" and all attributes and relationships the node involved with.\\
& 1. The scene shows a silver motorcycle standing on a paved ground.\\
& 2. Paved ground is typically suitable for riding.\\
& 3. Since bikes are usually used for riding or recreation, the setting suggests a riding activity. \\
& \textbf{Conclusion:} The man on the floor is most likely engaged in riding the motorcycle.\\[4pt]
\midrule

\textbf{SceneAlign overthink} 
& \textit{Change:} Randomly select and add a relationship (["building", "behind", "motorcycle"]) from the residual pool.\\
& 1. The scene shows a silver motorcycle standing on a paved ground, with a building located behind it.\\
& 2. A man is crouching on the ground, holding a white piece of paper and looking toward the motorcycle.\\
& 3. The explicit mention of the surrounding building introduces additional contextual structure, suggesting the motorcycle may be in a service, maintenance, or official facility area rather than a casual setting.\\
& 4. This additional environmental cue may bias the interpretation toward more formal or professional use of the motorcycle, rather than simple recreation.\\
& \textbf{Conclusion:} The man is most likely engaged in an inspection or diagnostic activity related to the motorcycle within a structured environment.\\[4pt]

\bottomrule 
\end{tabularx} 
\caption{\textbf{Case Study: Scene-Graph–Grounded Reasoning.} Structured comparison of grounded reasoning, showing how each perturbation (\textit{swap, replace, shorten, overthink}) leads to distinct failure modes while the positive CoT remains consistent with the scene graph.} 
\label{tab:case_study_scenealign} 
\end{table*}